\documentclass[letterpaper, 10 pt, conference]{ieeeconf}  

\IEEEoverridecommandlockouts                              

\usepackage{graphics} 
\usepackage{epsfig} 
\usepackage{mathptmx} 
\usepackage{times} 
\usepackage{amsmath} 
\usepackage{amssymb}  
\usepackage{booktabs}
\usepackage{dblfloatfix}
\usepackage{cancel}
\usepackage{hyperref}

\title{\LARGE \bf
When Absolute State Fails: Evaluating Proprioceptive Encodings for Robust Manipulation
}

\author{Maxime Alvarez$^{1}$$^{,}$$^{2}$, Ryo Watanabe$^{1}$, Paul Crook$^{1}$, Afshin Zeinaddini Meymand$^{1}$,\\ Suvin Kurian$^{1}$, Pablo Ferreiro$^{1}$, Genki Sano$^{1}$ 
\\
\thanks{$^{1}$ TELEXISTENCE Inc, Foundation Model Division, Japan.}
\thanks{$^{2}$ The University of Tokyo, Japan.}
}

\begin{document}

\maketitle
\thispagestyle{empty}
\pagestyle{empty}

\begin{abstract}

As end-to-end robotic policies are progressively deployed in the real world to solve real tasks, they face a gap between the training and inference conditions. Scaling the amount and diversity of the training data has shown some success in improving zero-shot generalization, yet robots still fail when faced with new, unseen test conditions. For instance, while robots with fixed frames of reference are common, those with moving frames pose a greater challenge for deployment. To address this specific instance of the issue, we present a study of strategies for encoding the robot's proprioceptive state to improve both in- and out-of-distribution performance at test time. Through a systematic study of joint representations, we find that a simple episode-wise relative frame provides the best trade-off between task performance and robustness, outperforming the baselines in extensive real-robot experiments conducted in a realistic test environment. The results suggest a practical path to leveraging data collected by robots with varying frames of reference and deployment to unseen test configurations.
\end{abstract}

\section{INTRODUCTION}

The transition of robotic manipulation from static, tightly controlled laboratory workcells to unstructured, dynamic environments represents one of the field's most pressing challenges. As robots are increasingly deployed in real-world settings, such as retail spaces, warehouses, and domestic environments, they are often equipped with mobile bases \cite{airoa-moma-2025, wu2024tidybot, autonomous-mobile-robot-intralogistics} or linear rail systems \cite{stretch, ghost} to extend their operational workspace. While imitation learning (IL) \cite{endtoend-deep-policy} has demonstrated remarkable success in teaching these robots complex tasks from human demonstrations, the policies inherently struggle to generalize when deployed under initial conditions that differ even slightly from their training data.

A critical but often overlooked source of this brittleness lies in how a robot’s own physical state is represented. When a robot operates from a fixed base, its internal coordinate frame aligns consistently with its workspace. However, when a robot features moving frames of reference, such as a manipulator mounted on a mobile carriage or linear rails, the absolute values of its proprioceptive state become highly variable across episodes. If an end-to-end policy is trained on absolute joint positions, an identical manipulation task executed from a starting position offset by just a few centimeters can result in catastrophic failure or erratic, unsafe movements.

\begin{figure}[!t]
    \includegraphics[width=0.5\textwidth]{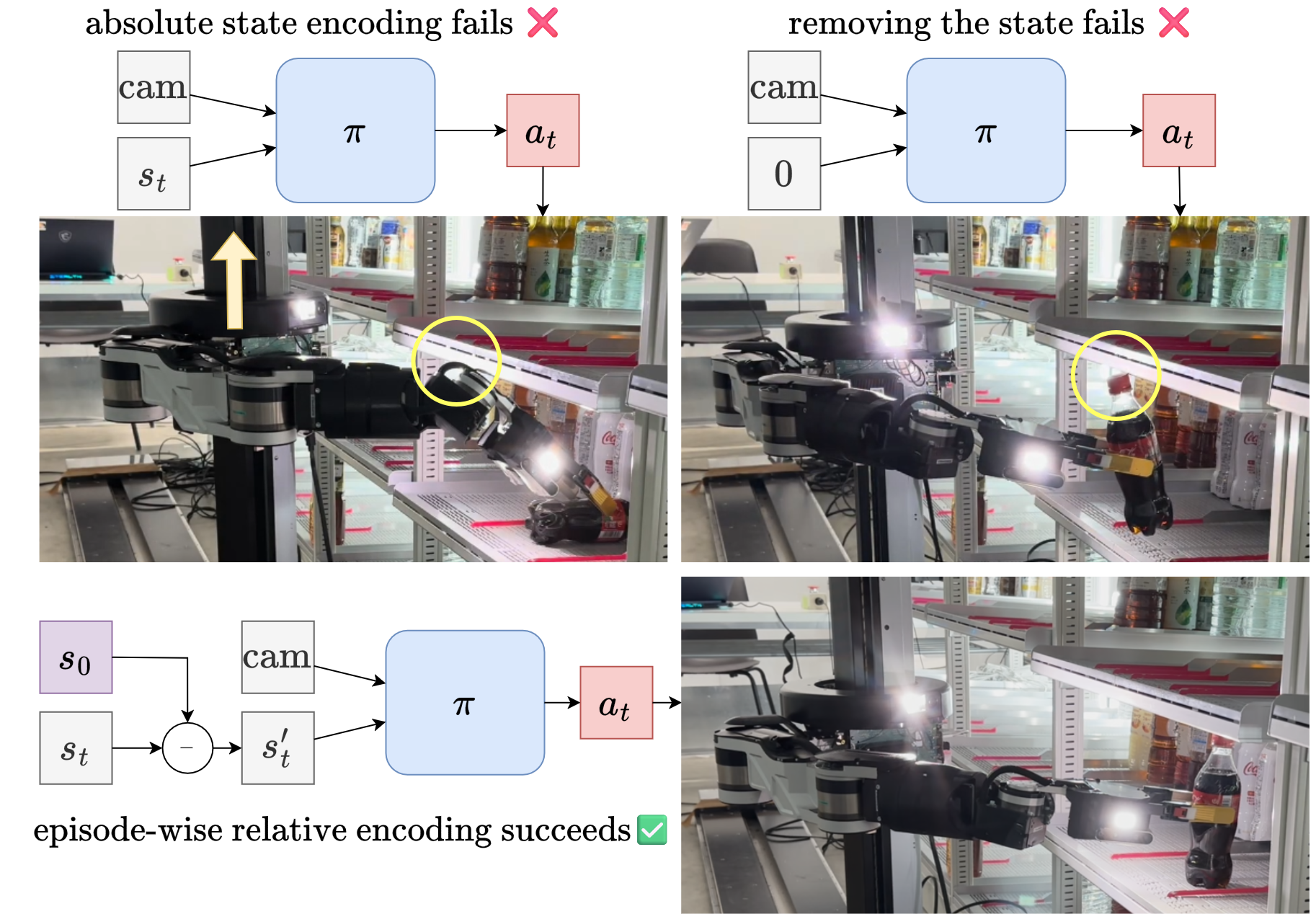}
    \caption{Comparison of the three state-action representations. Absolute state/action encoding fails and can produce unsafe behavior. Removing the state and using chunk-wise actions improves robustness but still suffers from drift, especially in precision-critical stages. The proposed episode-wise relative encoding strikes a balance between the two and successfully solves the task. In the top-left example, the wrist collides with the upper shelf, leading to failure. In the top-right example, the bottle tilts after collision. A video digest of our work is available here: \href{https://youtu.be/2tTUL_bIeW4}{https://youtu.be/2tTUL\_bIeW4}}
    \label{fig:main_fig}
\end{figure}

To date, the overwhelming majority of research aimed at closing this train-test distribution gap has focused on the visual domain. Massive strides have been made in extracting robust, invariant visual features \cite{kawaharazuka2025vla-survey, robotics11060139} ; however, proprioceptive inputs, such as joint positions and velocities, are still frequently fed into neural policies as raw, absolute numeric values. While some recent frameworks have begun exploring action spaces relative to chunk histories \cite{chi2024universal} or external object poses \cite{pmlr-v205-pan23a, zhu2022viola}, these methods either lose critical global state information or demand complex external pose estimators. The effectiveness of a purely intrinsic, robust proprioceptive encoding remains underexplored.

In this work, we address the specific challenge of proprioceptive drift and coordinate variability in mobile manipulation. We investigate alternative state- and action-representation strategies for a dual-rail robotic system tasked with precise visual manipulation. By moving away from absolute state encodings, we propose a simple yet highly effective episode-wise relative coordinate frame. This approach dynamically standardizes the robot's starting position at the start of each episode, effectively collapsing variance in initial conditions without sacrificing the fine-grained precision required for tasks such as grasping and object placement.

We evaluate our representations using an Action Chunking Transformer (ACT) \cite{zhao2023learning} architecture on a real-world bottle recovery task.

Our main contributions are as follows:
\begin{itemize}
    \item An investigation of three different state and action representations to study their impact on policy performance.
    \item Demonstration of significant improvement and robustness through extensive testing on a real robot in a realistic test environment.
    \item Identification that an episode-wise relative frame strikes a balance between generalizing to new states and exploiting state information, effectively collapsing the variance in starting conditions so the model learns more efficiently.
\end{itemize}

See Figure \ref{fig:main_fig} for an illustration of the representation strategies and instances of failures.

\section{RELATED WORKS}
As highlighted by \cite{lu2026visionproprioceptionpoliciesfailrobotic}, integrating proprioception information into vision-based policies can be difficult, and the literature is unclear about the benefits of doing so. As a consequence, various representations have emerged over time.

\textbf{Relative state representation.} Robot-frame relative actions are used in reinforcement learning. The robot defines an egocentric static frame, and actions are specified relative to that frame, and typically do not use world information, except in cases such as multi-agent \cite{multi-agent-lowe}. This is, for instance, the case in walking \cite{taskspace_bipedal}. In ZeroMimic \cite{shi2025zeromimic}, an ablation suggests that relative actions are superior to absolute actions in the camera frame. In UMI \cite{chi2024universal}, the authors define a relative frame using historical observations. In our case, we do not have a history of observations, so we replace the chunk-relative state with a constant 0 state and use a similar approach of chunk-relative actions.

\textbf{Object-oriented frame coordinates.} Some works use poses relative to an object in the world, which requires information outside of the robot \cite{pmlr-v205-pan23a, zhu2022viola}. Here, we use only the robot's intrinsic state.

\textbf{State-free policies.} By extending the visual observations available to the policy and using delta end-effector actions, State-Free Policies \cite{zhao2025nostate} have demonstrated stronger generalization in height and horizontal task placement variations. Although we use a state-free policy with chunk-wise delta actions as one of our baselines, we do not change the hardware to include additional vision inputs. In our task, the "full task observation" (as defined in \cite{zhao2025nostate}) is already available through the given visual inputs.
\section{METHOD}
\begin{figure}[h]
    \includegraphics[width=0.5\textwidth]{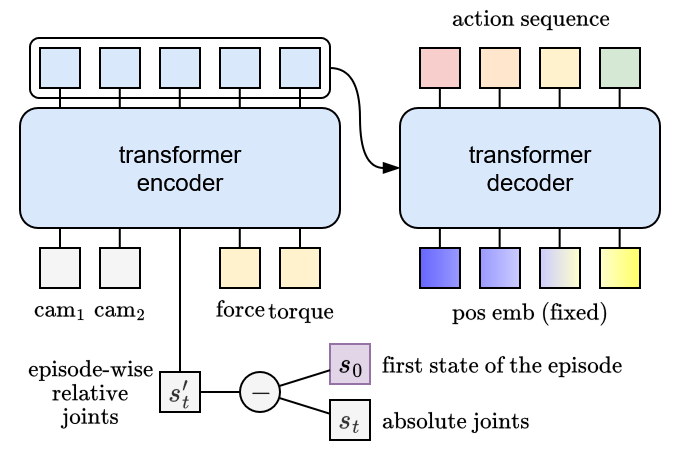}
    \caption{Illustration of the proposed episode-wise relative state and actions method applied to ACT.}
    \label{fig:proposed_method}
\end{figure}

To study the impact of state and action representations on policy performance in- and out-of-distribution, we evaluate three different state and action representations. The robot's state has 10 components, of which components 0 and 1 correspond to the linear rails controlling the position of the shoulder of the robot on the 2D plane facing the task environment. For simplicity, when referring to the state components 0 and 1 ($s_{\{0,1\}}$) we write $s$. The other components of the state are only normalized using the training dataset's statistics. For our robot control software, the actions are target joint commands, meaning that the actions are target states. The same components 0 and 1 in the actions refer to the 2D plane control, and we apply the same shortcut notation $a$ for $a_{\{0,1\}}$.
We compare:
\begin{itemize}
    \item \textbf{Absolute state and absolute actions}: the state is the current absolute joint values from the robot, and actions are the target absolute joint values sent to the robot system.
    \item \textbf{Episode-wise state and episode-wise actions}: at the beginning of each episode, the current absolute value of the state is defined as the origin, and all states and actions are encoded relatively to it. Essentially, this approach reduces to subtracting the state of each episode at time 0 from the state of the rest of the time steps.
    $$s^\prime_{i,t} = s_{i,t} - s_{i,0}$$
    $$a^\prime_{i,t} = a_{i,t} - s_{i,0}$$ where $i$ is the index of the episode, and $t$ is the timestep within the episode. See Figure \ref{fig:proposed_method} for an illustration of the episode-wise relative state.
    \item \textbf{No state and chunk-wise actions}: following UMI \cite{chi2024universal}, the state and actions are defined relative to the first state in the observation history. Given that in our setting we use a single time step of observation, the state is effectively 0. The actions are defined relative to the state at time t for the entire action chunk.
     $$s^\prime_{i,t} = \cancel{0}$$
     $$a^\prime_{i,t+k} = a_{i,t+k} - s_{i,t},\ \forall k \in [0, H]$$ where $H$ is the length of a single action chunk and $k$ is the index of the action within the chunk. This also an implementation of \cite{zhao2025nostate} in the context of our task.
\end{itemize}
For all other joints in the robot, the state and actions are represented in absolute values. We keep the other joints absolute as they are independent of the robot’s position on the rails at the start of the episode.

\section{EXPERIMENTS}
\subsection{Task}
\begin{figure*}[btp]
    \includegraphics[width=1.0\textwidth]{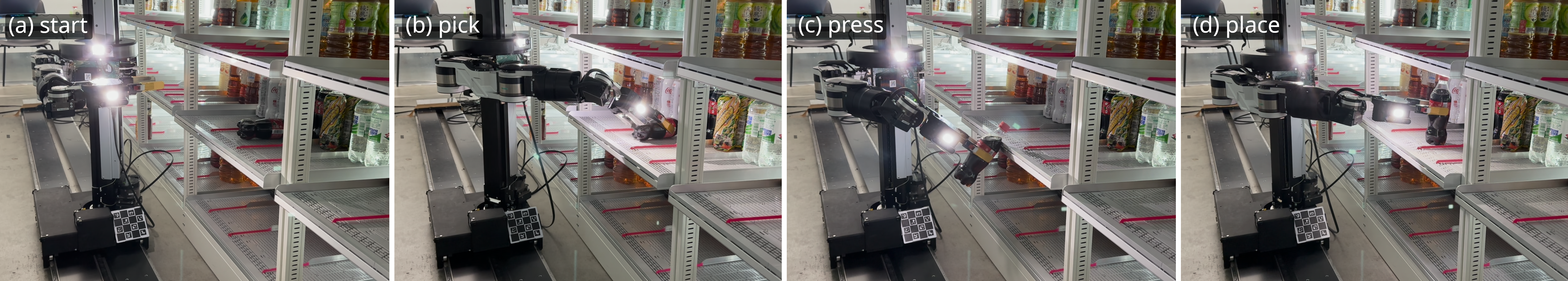}
    \caption{Bottle-recovery task decomposed into four stages: (a) \textit{Start}—initial state, (b) \textit{Pick}—approach and grasp the bottle, (c) \textit{Press}—move to the shelf edge and press the bottle against it, (d) \textit{Place}—move to the shelf and set the bottle upright.}
    \label{fig:task}
\end{figure*}

We reuse the bottle recovery task and the Ghost robot \cite{ghost} presented in FTACT \cite{watanabe2025ftactforcetorqueaware}, but extend the task from a single starting position in front of a table to a variable starting position at the back of a convenience store shelf. See Figure \ref{fig:task} for an illustration of the task.

\subsection{Model}
We use the ACT model \cite{zhao2023learning}, replacing the ResNet vision encoder with DINOv2-small \cite{oquab2023dinov2} and increasing the decoder's number of layers to 9. We use a cosine decay learning rate scheduler with a warmup phase of 1000 steps and a decay period of 150000 steps. We set the peak learning rate to 1.0e-4 and the decay learning rate to 1.0e-5. We keep the vision encoder frozen during training. We train on a single node with 8 NVIDIA H200 GPUs, using a batch size of 128 per GPU for a total of 200k gradient steps, with full FP32 precision.

\subsection{Dataset}
The dataset we use for this experiment has been collected by teleoperators solving the given task. There are 2500 episodes, totaling around 70h of data, at 50 Hz. The policy is fed three images: two images coming from the stereo cameras that form the robot's "head" and which are mounted on the arm carriage, and the third from a wrist camera mounted at the end of the arm. All images are resized to 224x224 to fit the pretrained vision encoder. The state has 10 dimensions: 2 dimensions are the X and Z rails, 7 dimensions correspond to the robot’s arm, and the last dimension is the gripper. To that state, we concatenate a 6D force-torque sensor calibrated to 0 at the beginning of the episode.

The data collection was done in 2 environments, A and B, and the evaluation was carried out in environment B only. The difference between the two environments lies in the shelf positions, bottle arrangements, and the distance of the robot rail from the base of the shelves, the latter varying by a few centimeters.

\begin{figure}[h]
    \includegraphics[width=0.5\textwidth]{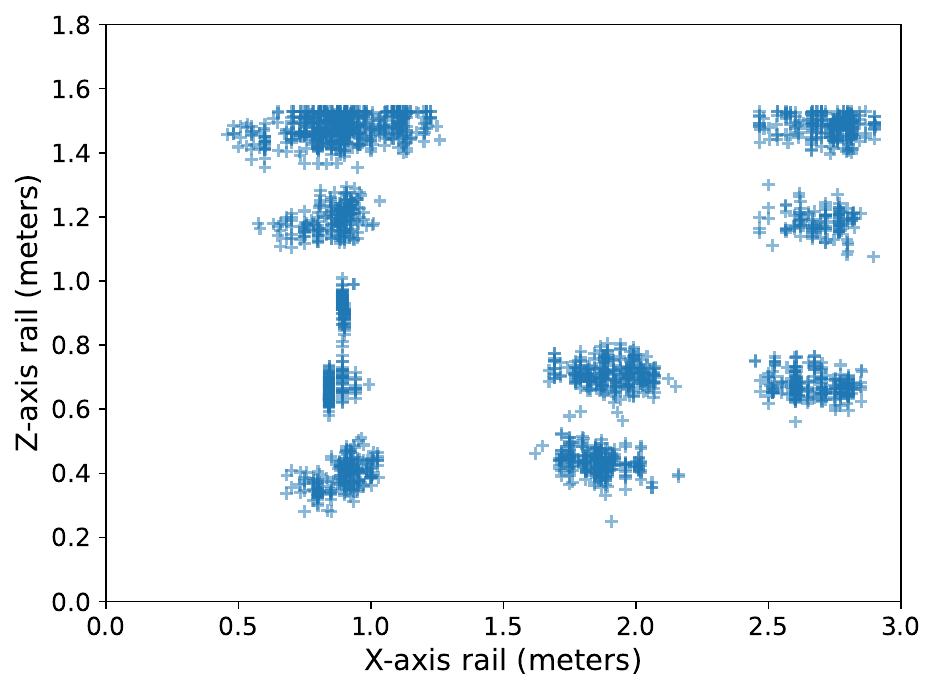}
    \caption{Distribution of the starting values for each episode in the dataset of the X (on the x-axis) and Z (on the y-axis) rails of the robot.}
    \label{fig:dataset_start_distribution}
\end{figure}

In Figure \ref{fig:dataset_start_distribution}, we show the starting position in X and Z positions for all episodes in the training dataset, when represented in absolute values.

\subsection{Evaluation}
We evaluate the 3 policies in-distribution and out-of-distribution, defining out-of-distribution as any X/Z position unseen in the training dataset and unseen bottle types. In total, we evaluate 6 Z values (4 for the in-distribution evaluations and 2 for the out-of-distribution) and 36 X values. The changes in Z values are significant (from one shelf to another) while the changes in X values are small (bottles being next to each other). For the out-of-distribution X-values, we use a different cabinet. See the Appendix for an illustration of the test environment with the robot.

We score each episode on a 0 to 4 scale as such:
\begin{itemize}
    \item 0 when the robot cannot grasp the bottle
    \item +1 when the robot can grasp the bottle
    \item +1 when the robot can upright the bottle
    \item +1 when the robot puts the bottom of the bottle on the shelf
    \item +1 when the robot opens the gripper and the bottle is stable (does not fall over)
\end{itemize}
This leads to a total of 4 points when the episode is successful. Each episode has a timeout of 5 minutes, beyond which the episode is halted, and the score is tallied.

\section{RESULTS}
\begin{figure}[h]
    \includegraphics[width=0.5\textwidth]{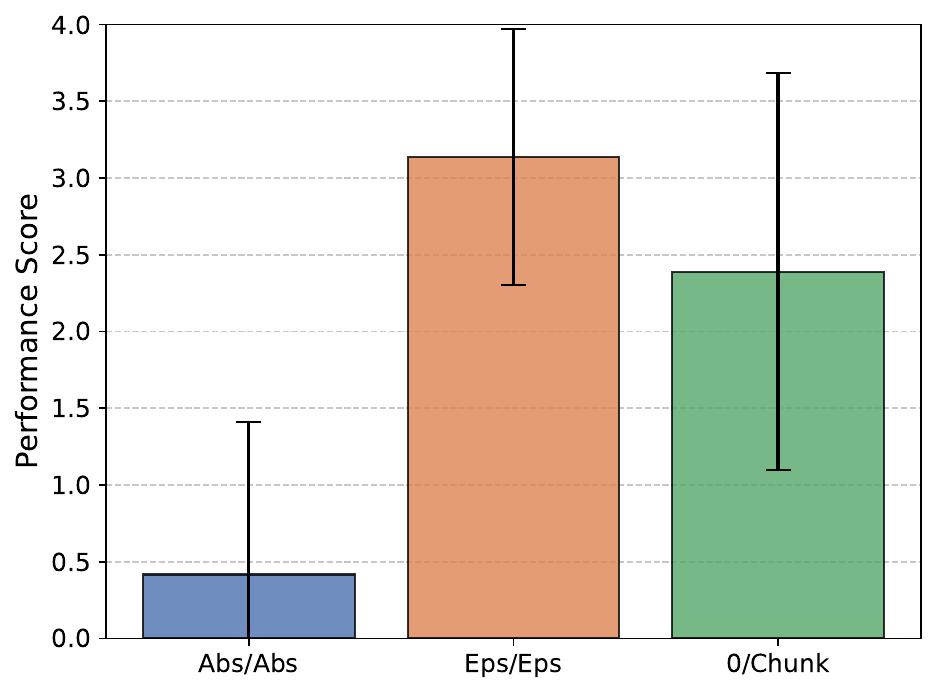}
    \caption{The \textbf{episode-wise state and episode-wise actions} (Eps/Eps) performs the best, with \textbf{no state and chunk-wise actions} (0/Chunk) performing slightly lower and with a much higher variance, while \textbf{absolute state and absolute actions} (Abs/Abs) fails completely.}
    \label{fig:plot_results}
\end{figure}

In Table \ref{tab:numerical_results} and Figure \ref{fig:plot_results}, we present the results from the evaluation for in-distribution and out-of-distribution as well as the average of both. We report the mean score with the standard deviation of the score. In Table \ref{tab:success_rate_results}, we present the success rate for a binary success/failure scoring of the same evaluation episodes.

\begin{table}[h]
    \centering
    \caption{Average scores (mean \textpm\ SD) for ID and OOD evaluations.}
    \label{tab:numerical_results}
    \begin{tabular}{c c c c}
         \toprule
         & \textbf{Abs/Abs} & \textbf{Eps/Eps} & \textbf{0/Chunk}  \\
         \midrule
         In-distribution & 0.75 \textpm 1.25 & \textbf{3.30 \textpm 0.80} & 2.25 \textpm 1.52 \\
         Out-of-distribution & 0.00 \textpm 0.00 & \textbf{2.94 \textpm 0.85} & 2.56 \textpm 0.96 \\
         Total & 0.38 \textpm 0.99 & \textbf{3.12 \textpm 0.83} & 2.41 \textpm 1.29 \\
         \bottomrule
    \end{tabular}
\end{table}

\begin{table}[h]
    \centering
    \caption{Success rate for ID and OOD evaluations.}
    \label{tab:success_rate_results}
    \begin{tabular}{c c c c}
         \toprule
         & \textbf{Abs/Abs} & \textbf{Eps/Eps} & \textbf{0/Chunk}  \\
         \midrule
         In-distribution & 5.0\% & \textbf{50.0}\% & 20.0\% \\
         Out-of-distribution & 0.0\% & \textbf{25.0}\% & 20.0\% \\
         Total & 2.5\% & \textbf{37.5}\% & 20.0\% \\
         \bottomrule
    \end{tabular}
\end{table}

The \textbf{absolute state and absolute actions} setting (noted Abs/Abs) performs the worst. In distribution, the policy failed consistently to grasp the bottle and succeeded in grasping it 6 times, and managed to fully complete the task only once. During the out-of-distribution evaluation, the policy displayed dangerous movements, prompting us to halt the evaluation and give a 0 score for out-of-distribution.

The \textbf{no state and chunk-wise actions} setting (noted 0/Chunk)  performs poorly, completely failing to grasp the bottle in 5 out of 36 cases. Interestingly, this policy does not suffer any performance loss from the out-of-distribution cases, even improving the average score (+13\%). The average score standard deviation is higher than all other policies, showing unreliable performance.

The \textbf{episode-wise state and episode-wise actions} setting (noted Eps/Eps) performs the best overall but suffers from some performance degradation (-11\%) in the out-of-distribution setting. This setting shows the best average performance and seems to strike a compromise between generalizing to new states and exploiting the state information for improved performance in the in-distribution setting. As detailed in Appendix A, this is likely because the episode-wise relative representation effectively collapses the variance in starting conditions, allowing the model to learn more efficiently across all training episodes.

\section{CONCLUSION}
In this work, we studied 3 different approaches to encoding the state and actions of linear joints and found that while not inputting the state at all (0/Chunk setting) leads to less performance drop when in the OOD setting, its performance in the ID setting is lower than that of relative episode-wise state and actions. Episode-wise state and actions get the best performance overall at the price of a drop in average score when in the OOD setting.

While this paper studies a specific task and a specific robot, the results are expected to hold with other robots in other settings that also have linear joints, such as the Agitbot G1, where the torso is set on a vertical rail.

The current methods still suffer from generalization issues when out of distribution, as is common for imitation learning-based methods. Also, we leave for future research the study of the case of non-linear joints, as the episode-wise approach might not be applicable.

\addtolength{\textheight}{-12cm}   
\bibliographystyle{IEEEtran}
\bibliography{main}
\newpage
\section*{APPENDIX}
\subsection{Environment}
\begin{figure}[h]
    \centering
    \includegraphics[width=0.8\linewidth]{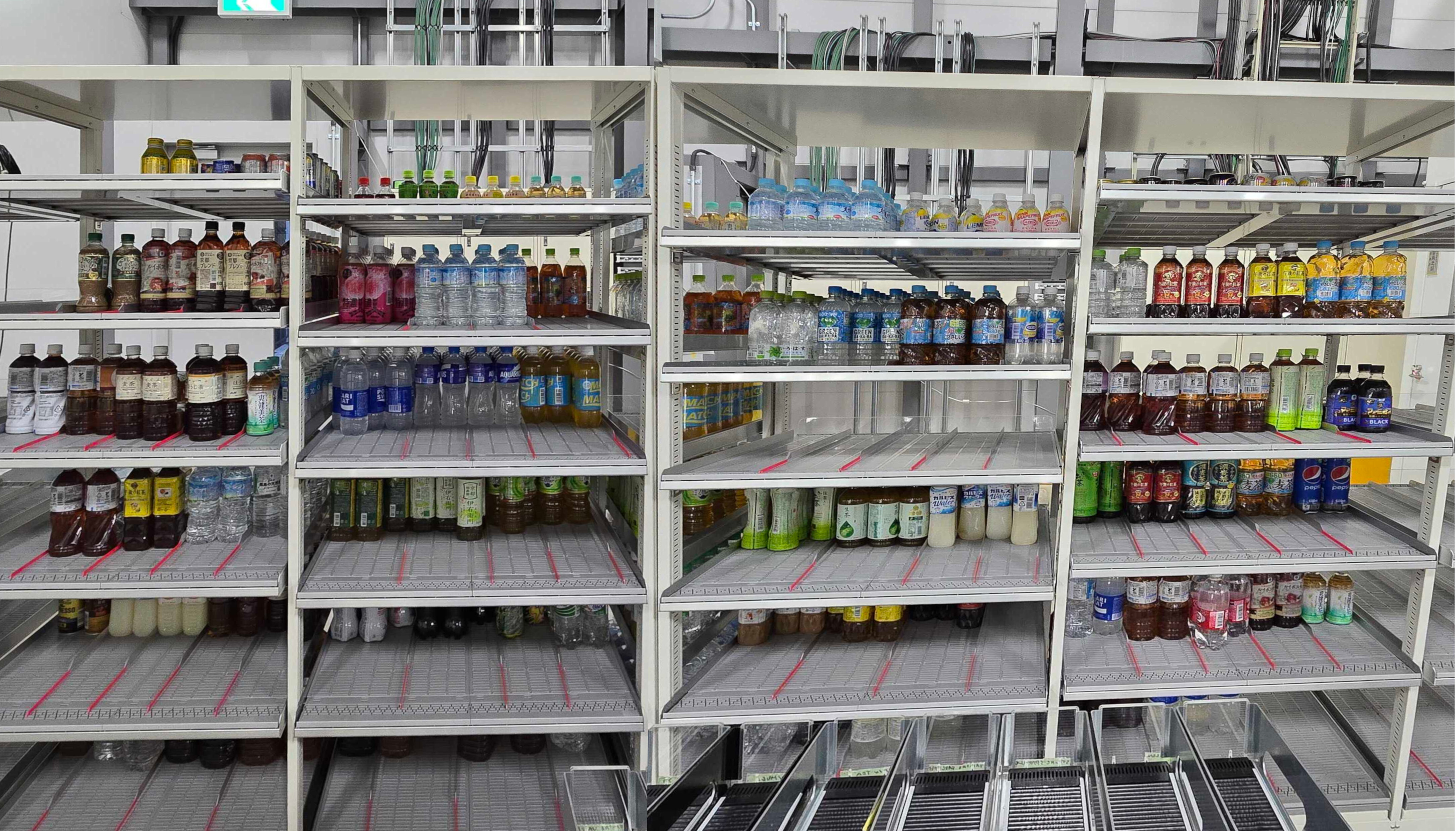}
    \caption{Stitched pictures of the test environment. Multiple shelves spread across multiple cabinets offering a variety of initial conditions for the X and Z rails.}
    \label{fig:cabinet}
\end{figure}

In Figure \ref{fig:cabinet}, we show the environment in which the data was collected and in which the evaluation takes place. The environment is made up of multiple cabinets, each with multiple levels. Out-of-distribution is defined as cabinet and level combinations unseen in the dataset.
\subsection{Dataset distribution}
\begin{figure}
    \centering
    \includegraphics[width=0.8\linewidth]{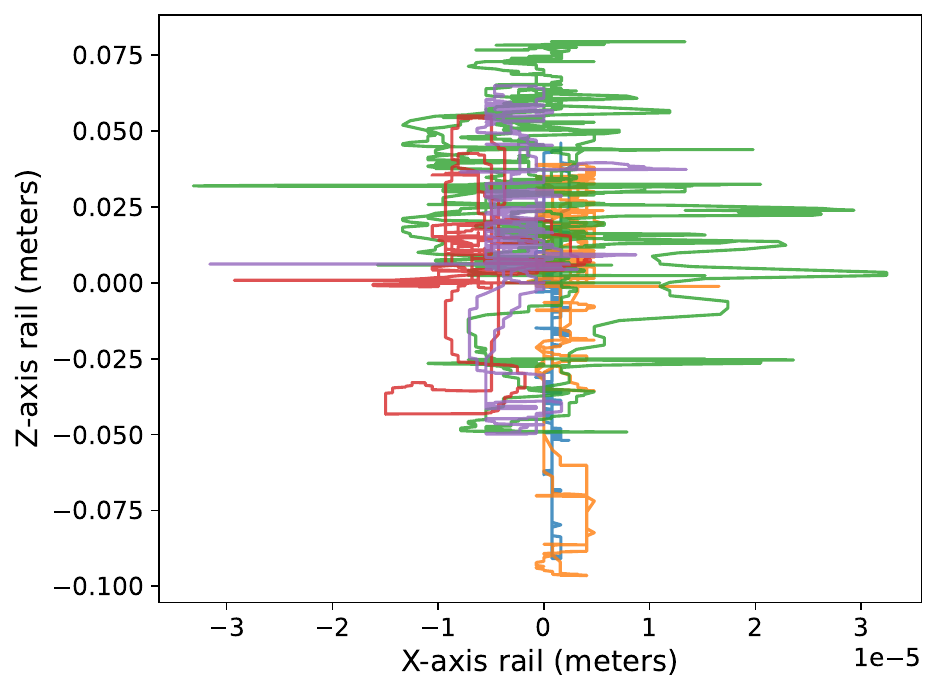}
    \caption{X and Z values for 5 random episodes from the training dataset, represented in episode-wise relative values.}
    \label{fig:sample_trajectories}
\end{figure}

In Figure \ref{fig:sample_trajectories}, we show 5 random trajectories represented in episode-wise relative coordinates. We can see that the Z-axis has some variance intra-episode, while the X-axis stays almost fixed for the duration of the whole episode, as the scale of the X-axis shows ($1\times 10^{-5}$). The episode-wise relative representation collapses the starting conditions, allowing the model to learn more efficiently from all episodes.
\subsection{Videos}
We provide two videos, one overview of this work: \href{https://youtu.be/2tTUL_bIeW4}{https://youtu.be/2tTUL\_bIeW4}, and one $\times 1.5$ accelerated video of the Eps/Eps configuration solving the task: \href{https://youtu.be/Fq9ufXxWj20}{https://youtu.be/Fq9ufXxWj20}.

\end{document}